\documentclass[pdflatex,sn-mathphys-num]{sn-jnl}% Math and Physical Sciences Numbered Reference Style
\usepackage{graphicx}%
\usepackage{multirow}%
\usepackage{amsmath,amssymb,amsfonts}%
\usepackage{amsthm}%
\usepackage{mathrsfs}%
\usepackage[title]{appendix}%
\usepackage{xcolor}%
\usepackage{textcomp}%
\usepackage{manyfoot}%
\usepackage{booktabs}%
\usepackage{algorithm}%
\usepackage{algorithmicx}%
\usepackage{algpseudocode}%
\usepackage{listings}%
\usepackage{xcolor}%
\usepackage{algpseudocode} % 

\theoremstyle{thmstyleone}%
\theoremstyle{thmstyletwo}%

\theoremstyle{thmstylethree}%

\begin{document}

\title[Article Title]{Unsupervised Mapping of Urban Tree Diversity using Spatially-aware Visual Clustering}


\author*[1,2]{\fnm{Diaa Addeen} \sur{ Abuhani}}\email{diaa@mit.edu}

\author[1,3]{\fnm{Marco} \sur{Seccaroni}}\email{marco.seccaroni@polimi.it}

\author*[1]{\fnm{Martina} \sur{Mazzarello}}\email{mmazz@mit.edu}

\author[2]{\fnm{Imran} \sur{Zualkernan}}\email{izualkernan@aus.edu}

\author[1]{\fnm{Fabio} \sur{Duarte}}\email{fduarte@mit.edu}

\author[1,3]{\fnm{Carlo} \sur{Ratti}}\email{ratti@mit.edu}

\affil*[1]{\orgdiv{Department of Urban Studies and Planning, Senseable City Laboratory}, \orgname{Massachusetts Institute of Technology}, \orgaddress{\street{77 Massachusetts Ave}, \city{Cambridge}, \postcode{MA 02139}, \state{Massachusetts}, \country{USA}}}

\affil[2]{\orgdiv{Department of Computer Science and Engineering}, \orgname{American University of Sharjah}, \orgaddress{\street{127 Street}, \city{University City}, \postcode{26666}, \state{Sharjah}, \country{UAE}}}

\affil[3]{\orgdiv{Department of Architecture, Built Environment, and Construction Engineering}, \orgname{Politecnico di Milano}, \orgaddress{\street{ Piazza Leonardo da Vinci, 32}, \city{Milano}, \postcode{20133}, \state{Milan}, \country{Italy}}}


\abstract{Urban tree biodiversity is critical for climate resilience, ecological stability, and livability in cities, yet most municipalities lack detailed knowledge of their canopies. Field-based inventories provide reliable estimates of Shannon and Simpson diversity but are costly and time-consuming, while supervised AI methods require labeled data that often fail to generalize across regions. We introduce an unsupervised clustering framework that integrates visual embeddings from street-level imagery with spatial planting patterns to estimate biodiversity without labels. Applied to eight North American cities, the method recovers genus-level diversity patterns with high fidelity, achieving low Wasserstein distances to ground truth for Shannon and Simpson indices and preserving spatial autocorrelation. This scalable, fine-grained approach enables biodiversity mapping in cities lacking detailed inventories and offers a pathway for continuous, low-cost monitoring to support equitable access to greenery and adaptive management of urban ecosystems.}

\keywords{Biodiversity, Unsupervised Learning, Clustering, Streetviews}



\maketitle

\section{Introduction}\label{sec1}

Quantifying biodiversity in urban forests is essential to assess ecosystem health and to guide evidence-based management and policy decisions for urban green spaces. Widely used metrics like Shannon entropy and Simpson’s index integrate both richness (number of taxa) and evenness (distribution of abundances) into a single value, providing interpretable indicators for comparing ecological conditions across locations and over time \cite{magurran_diversity_1988, magurran_choosing_1988}. However, accurate estimation of these indices at the genus or species level typically requires detailed and geo-referenced taxonomic inventories, most often collected through arborist-led field surveys \cite{nielsen2014review}. Although these surveys are reliable, they are resource intensive. Urban forestry programs may spend up to \$65 per tree annually on inventory, monitoring, and management, which has limited large-scale implementation to only a few well-funded cities \cite{davies2011mapping, escobedo_costs_2009}.

In the past 50 years, adapting to climate change has become a priority for cities around the world. Seminal and contemporary studies \cite{boonman_more_2024, cardinale_biodiversity_2012, jones_scale_2024, aronson_global_2014} highlight that anthropogenic changes in the past few decades have caused a rapid loss of urban biodiversity. In addition to supporting biodiversity and urban ecosystems \cite{helden_urban_2012}, trees play a critical role in maintaining livability by contributing to equitable access to greenery \cite{gunnarsson_effects_2017, lin_opportunity_2014, liu_are_2022}, improved air quality \cite{hewitt_using_2020, mcpherson_structure_2016}, and thermal comfort \cite{wang_tree_2021, ishimatsu_developing_2021}. Yet, tree planting strategies are often guided by aesthetic preferences rather than ecological function \cite{zhao_visual_2017}, resulting in reduced diversity and increased vulnerability to pests, diseases, and climate stressors \cite{liu_are_2022}. A more diverse and ecologically informed urban canopy can buffer against these risks and adapt over time \cite{muluneh_contributions_2022, esperon-rodriguez_climate_2022, jenerette_climate_2016}.

Open access data sources, such as Street View Imagery (SVI), offer a scalable alternative for biodiversity monitoring due to their extensive spatial coverage \cite{beery_where_2023}. However, SVI datasets are unlabeled, meaning the images do not specify the tree species present and require substantial pre-processing to extract ecological information suitable for diversity estimations. Identifying individual taxa from imagery is particularly challenging: trees of the same species may look very different depending on season, age, or management, while distinct species can appear deceptively similar. These difficulties make automated classification highly sensitive to subtle morphological cues. Current supervised ecological AI models \cite{stevens_bioclip_2024, sastry2024taxabindunifiedembeddingspace}, while effective within their training domains, often underperform when applied to regions or taxa not well represented in training data, largely due to geographic domain shift \cite{sierra2025divshift}. 

Self-supervised learning offers a promising alternative by eliminating the need for labeled data \cite{chen_simple_2020}. However, when applied to limited or homogeneous datasets, these methods often under-perform due to overfitting and poor generalization, as robust representations typically require large and diverse training samples. Few clustering approaches have been explicitly designed to preserve abundance distributions in a way that enables accurate Shannon and Simpson estimation from unlabeled imagery. This capability is essential for producing ecologically meaningful biodiversity metrics from large-scale, unlabeled datasets.

In this work, we address the challenge of large-scale, label-free biodiversity monitoring by developing an unsupervised clustering approach that estimates genus-level urban tree diversity directly from street-level imagery. We apply the framework across eight North American cities with contrasting forest compositions and imbalance, and evaluate its ability to recover biodiversity metrics such as Shannon and Simpson indices as well as spatial distribution patterns. While our evaluation is necessarily constrained to cities with available ground-truth inventories, the results demonstrate that the approach generalizes across diverse urban forest contexts and provides a scalable pathway for integrating biodiversity insights into urban resilience planning.

\section{Results}\label{sec2}

We evaluated our framework in eight North American cities spanning a wide range of forest compositions and genera imbalances (see Section~\ref{Data}). The method consistently recovered meaningful biodiversity patterns, demonstrating robustness even in cities dominated by a few genera. Table~\ref{tab:city_metrics} reports results for the best-performing configuration, selected through hyperparameter tuning (see Appendix~\ref{appendix:hyperparam}) and ablation study (see Appendix~\ref{ablation}), which achieved a balanced trade-off between richness accuracy, clustering quality, and diversity preservation.

\begin{table}[ht]
\centering
\caption{Clustering performance for our clustering framework across cities. 
Metrics include richness error ($RMSE_{\alpha} \downarrow$), clustering quality (V-score $\uparrow$), 
and distributional agreement with ground-truth Shannon (scaled) and Simpson diversity 
measured using the 1-Wasserstein distance ($W_1 \downarrow$). 
Values are reported as mean [95\% CI] through bootstrapping \cite{justus2024bootstrapconfidenceintervalscomparative}.}
\label{tab:city_metrics}
\begin{tabular}{lcccc}
\toprule
City & $RMSE_{\alpha} \downarrow$ & V-score $\uparrow$ & $W_{1,\text{Shannon}} \downarrow$ & $W_{1,\text{Simpson}} \downarrow$ \\
\midrule
Calgary        & 4.39 [4.20--4.59] & 0.397 [0.387--0.406] & 0.215 [0.208--0.222] & 0.178 [0.169--0.186] \\
New York       & 3.42 [3.34--3.51] & 0.476 [0.471--0.480] & 0.073 [0.070--0.077] & 0.064 [0.059--0.068] \\
Columbus       & 5.99 [5.76--6.22] & 0.430 [0.422--0.437] & 0.177 [0.172--0.183] & 0.140 [0.134--0.147] \\
Denver         & 5.39 [5.15--5.62] & 0.487 [0.480--0.494] & 0.069 [0.065--0.073] & 0.040 [0.036--0.045] \\
Los Angeles    & 3.74 [3.63--3.84] & 0.477 [0.473--0.481] & 0.070 [0.067--0.073] & 0.067 [0.064--0.071] \\
Seattle        & 9.94 [9.56--10.3] & 0.487 [0.481--0.493] & 0.155 [0.150--0.160] & 0.095 [0.089--0.101] \\
Washington& 15.6 [14.7--16.3] & 0.498 [0.491--0.505] & 0.133 [0.127--0.140] & 0.033 [0.029--0.039] \\
San Francisco  & 10.0 [9.17--10.8] & 0.493 [0.485--0.501] & 0.034 [0.028--0.041] & 0.028 [0.023--0.034] \\
\bottomrule
\end{tabular}
\end{table}

\paragraph{Richness and V-score results.}
As shown in Table~\ref{tab:city_metrics}, genera imbalance strongly influenced richness estimation error ($RMSE_{\alpha}$), though not uniformly across cities. In highly uneven forests such as Washington and Seattle, $RMSE_{\alpha}$ exceeded within-city variability, underscoring the difficulty of estimating richness under strong dominance patterns. By contrast, in more balanced systems such as Los Angeles, Denver, and New York, errors were smaller than or comparable to observed standard deviations, and in several cases outperformed a naïve mean-predictor baseline (see Appendix~\ref{RichnessSD}). Despite these differences, the V-score remained consistent across all datasets (0.397–0.498), indicating that clustering preserved the overall partitioning structure even when richness estimates deviated. These findings suggest that richness is not the primary strength of our approach, but performance is competitive in balanced forests and improves over statistical baselines.

\paragraph{Diversity results.}
For Shannon and Simpson diversity, we assessed agreement with ground truth using the Wasserstein distance \cite{Panaretos_2019}. This metric quantifies the minimum "cost" of transforming one distribution into another, accounting not only for absolute differences but also for their positions across the value range. Unlike RMSE, which captures pointwise error, the Wasserstein distance is sensitive to systematic shifts in distributional shape and spread, making it particularly suitable for ecological indices.

\begin{figure}
    \centering
    \includegraphics[width=\linewidth]{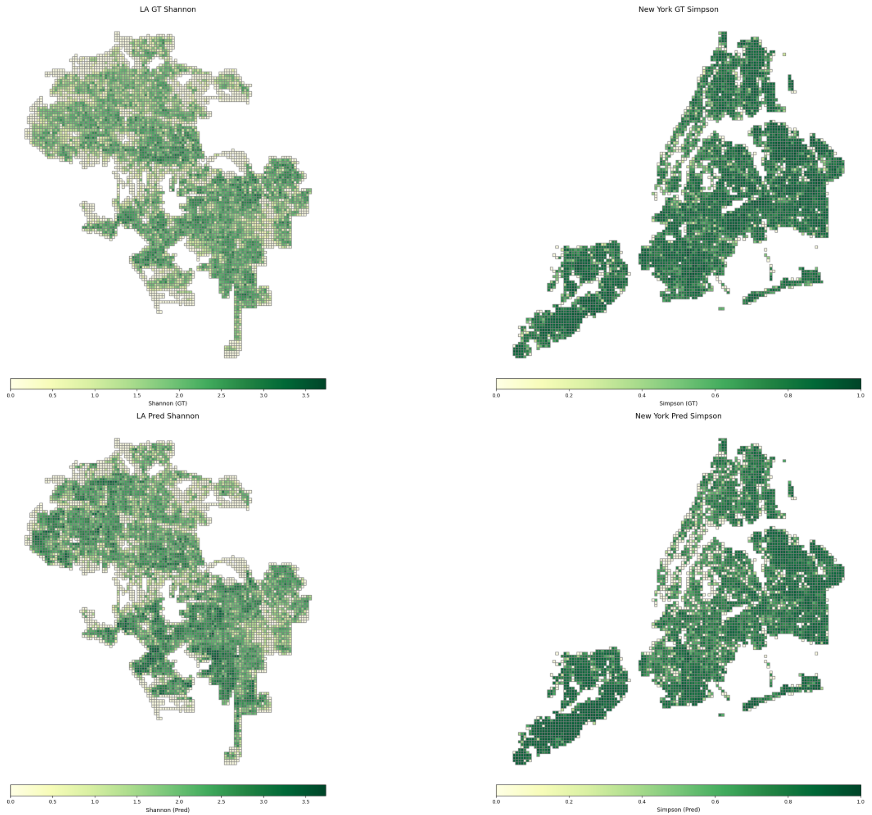}
    \caption{A comparison between ground truth and predicted values of Shannon entropy in L.A. and Simpson entropy in New York.}
    \label{fig:Results}
\end{figure}

In our datasets, Simpson diversity is bounded between 0 and 1, while Shannon diversity can reach values of about 3.5 (see Section~\ref{Data}). To enable direct comparison across indices and cities, Shannon values were normalized by their relative maximum, placing both metrics on a common [0,1] scale. In this context, Wasserstein distances below 0.05 indicate very strong alignment with ground truth distributions, while values near 0.1 represent moderate but still ecologically meaningful agreement \cite{panaretos2020invitation}. Cities such as Washington (0.133 for Shannon, 0.033 for Simpson) and San Francisco (0.034 for Shannon, 0.028 for Simpson) demonstrated strong alignment for both indices, with proportionally smaller errors for Simpson. Calgary, by contrast, exhibited the weakest alignment across metrics. We hypothesize that this may be linked to more aggressive visual obfuscation apparent in Canadian Street View imagery compared to the United States, where stronger privacy safeguards and stricter blurring protocols are enforced \cite{murphy_mapping}. Fig.~\ref{fig:Results} illustrates this by comparing predicted and observed values in Los Angeles (Shannon) and New York (Simpson). A full set of city-level results, including baseline comparisons and variability relative to observed standard deviations, is provided in Appendix~\ref{RichnessSD}, which contextualizes how performance varies across urban forest structures.

\paragraph{Diagnosis of possible entropy inflation.}
One potential concern with unsupervised clustering is that lower distributional errors may arise not from genuine recovery of community structure but from artificial entropy inflation, e.g., through oversplitting or the proliferation of singleton clusters. To assess this, we performed a dedicated diagnostic analysis (Appendix~\ref{diagnosis}) that examined oversplitting factors, purity, evenness shifts, and rank--abundance correlations between predicted and observed communities against a supervised model using linear probing on BioCLIP \cite{stevens_bioclip_2024} model. Across all cities, the singleton ratio was zero, oversplitting factors remained close to one, and evenness shifts were modest, with New York even showing a slight decrease. At the same time, rank--abundance correlations were consistently high ($\rho \approx 0.9$), indicating that predicted clusters preserved the overall shape of community distributions. These findings confirm that the improved alignment of Shannon and Simpson diversity under the unsupervised approach is not an artifact of inflated entropy, but rather reflects ecologically meaningful recovery of biodiversity structure.

Overall, the method achieved low Wasserstein distances for both Shannon and Simpson diversity across diverse urban forest conditions, including in highly imbalanced contexts. High Simpson accuracy suggests the framework effectively captured dominant genera distributions, while strong Shannon performance indicates robustness to abundance evenness patterns. Taken together, these results demonstrate that our approach can accurately recover both abundance distributions and spatial diversity patterns with strong spatial fidelity (see Appendix~\ref{spatial}) from entirely unlabeled street-level imagery, directly addressing the lack of scalable, label-free biodiversity mapping methods highlighted in the Introduction.

\section{Discussion}\label{sec3}
Our results demonstrate that an unsupervised and spatially aware clustering framework can recover both magnitude and spatially weighted biodiversity measures from unlabeled street-level imagery with high fidelity, even under markedly imbalanced genera distributions. Across diverse urban contexts, the method achieved low Wasserstein distances for Shannon and Simpson diversity, while preserving spatial autocorrelation patterns with minimal deviation from ground truth. These findings are consistent with previous reports that dominant taxa patterns are easier to recover than fine-scale evenness \cite{magurran_diversity_1988, beery_where_2023}, but extend prior work by showing that evenness-sensitive indices such as Shannon can also be estimated reliably when spatial priors are incorporated. Crucially, by enabling robust, label-free biodiversity mapping, this approach offers a scalable tool that could be integrated into municipal monitoring systems, guide nature-based policy, and support adaptive planning in cities with limited ecological data infrastructure.

Our ablation analysis (Appendix~\ref{ablation}) showed that the joint use of outlier elimination and cluster merging provided the best balance between clustering quality, richness accuracy, and preservation of diversity structure. This result indicates that addressing noise (through elimination) and fragmentation (through merging) in tandem is particularly effective for recovering meaningful ecological patterns from unlabeled imagery. More broadly, it suggests that unsupervised clustering in environmental applications benefits from strategies that explicitly mitigate both spurious assignments and over-segmentation. While prior studies have applied label-free clustering to infer urban landforms from very high-resolution satellite data \cite{metzler2025unsupervised} or crop types from multispectral time series \cite{wang_crop_2019}, such approaches typically rely on costly or proprietary datasets. By contrast, our framework demonstrates that freely available street-level imagery can support scalable, label-free biodiversity assessment in urban forests, extending applicability to regions where high-resolution remote sensing is not accessible.

From an applied perspective, the framework offers a scalable alternative to field-based inventories for monitoring urban tree diversity and can be integrated into urban forestry programs. Spatially explicit Shannon and Simpson maps at the city scale can guide planting strategies, highlight areas at risk of monocultures, and support temporal monitoring, while reducing the effort and cost of repeated ground surveys. Importantly, this directly addresses the central gap identified in the Introduction in regards to the inability of existing methods to recover abundance distributions and spatial diversity patterns from unlabeled imagery. By overcoming this limitation, our approach provides a pathway for continuous, cost-effective biodiversity monitoring that can inform urban resilience planning.

Several constraints remain. Street-level imagery is updated at irregular intervals, sometimes years apart, which complicates the interpretation of temporal change. In practice, this means the framework is better suited to structural baselines and cross-sectional monitoring than to precise year-to-year trend detection. Coverage is uneven globally, with gaps in rural areas, private developments, and in certain countries; thus, results are most applicable in dense urban cores of cities with established GSV infrastructure, and less transferable to places where coverage is patchy. Privacy-driven blurring can obscure morphological features needed for genus discrimination, potentially biasing species detection toward larger or more distinctive trees, and under-representing genera with finer-grained leaf or bark traits. Seasonal foliage changes, variable lighting, and transient occlusions (e.g. vehicles or pedestrians) introduce visual noise, which can reduce consistency across time and space and necessitate larger sample sizes to stabilize estimates. Additionally, the method only captures the visible streetscape; off-street vegetation is excluded, which can underestimate diversity in areas where backyards, courtyards, or parks hold a significant share of urban trees. Taken together, these factors highlight that while the method is effective for relative comparisons across cities and neighborhoods, it should be complemented with other data sources for absolute inventories or long-term ecological baselining.

Looking ahead, performance could be enhanced by incorporating vision–language models (VLMs) that leverage fine-grained textual descriptors of morphological traits \cite{saha2024improved}, improving discrimination between visually similar trees. As model efficiency advances, large-scale VLM integration may become feasible. Further gains may come from foundational models trained on ecologically diverse, high-resolution vegetation datasets, coupled with domain-adaptive fine tuning to improve generalization to underrepresented regions. In parallel, prior work shows that satellite RGB imagery alone is insufficient for reliable urban tree assessment \cite{beery_auto_2022, wegner_cataloging_2016}; integrating complementary modalities such as airborne LiDAR for structural detail and high-resolution multispectral data or recent foundation models \cite{brown2025alphaearthfoundationsembeddingfield} for spectral cues would enrich representation and enable a more holistic view of urban forest diversity. Beyond technical gains, these enhancements would directly support cross-city generalization, allowing models to adapt to ecological and urban form variability across different contexts. Moreover, the fusion of tree diversity indicators with other environmental datasets such as air quality, temperature, or land use would situate this work within the broader trend of multimodal urban environmental monitoring, creating synergies for applications in climate resilience, ecosystem services evaluation, and nature-based urban planning.

In sum, this work establishes that unsupervised clustering of spatial and visual embeddings can be a practical, transferable, and data-efficient approach to quantifying urban tree diversity. By recovering both the magnitude and spatial organization of Shannon and Simpson diversity indices, the framework offers a pathway for continuous, city-scale biodiversity monitoring and supports the development of more resilient, ecologically informed urban forests. By enabling the recovery of both abundance distributions and spatial diversity patterns from unlabeled imagery, our approach opens the door to scalable, globally comparable biodiversity mapping for urban environments.

\section{Methods}\label{sec4}
\begin{figure}[h]
    \centering
    \includegraphics[width=\linewidth]{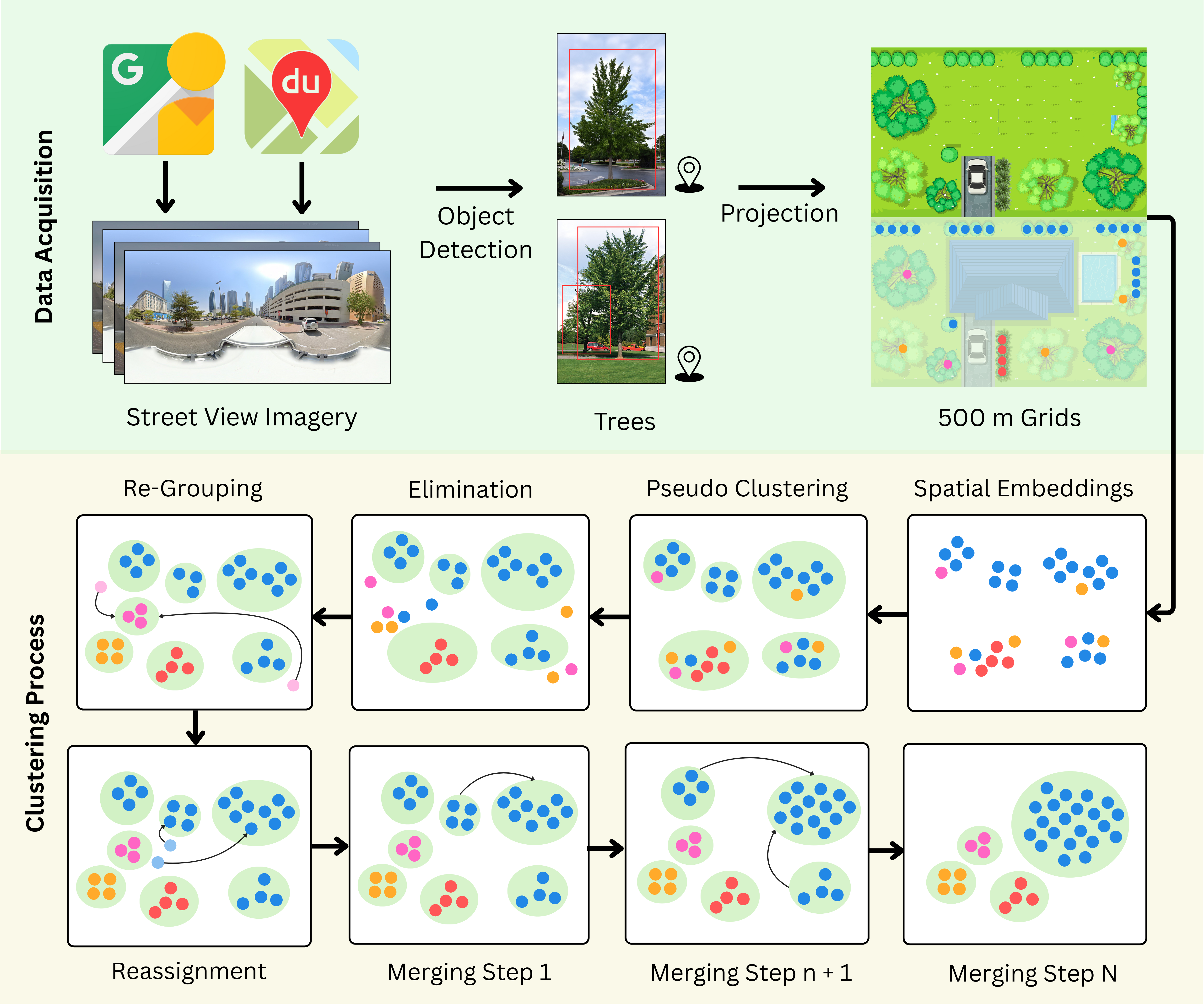}
    \caption{Overview of the methodology. Street-level imagery (see Section~\ref{Data}) is processed with an object detection model in prior work to extract tree instances, which are geo-located and mapped into 500 m grids. In our framework, spatial embeddings are clustered through an iterative procedure involving pseudo-clustering, merging, elimination, reassignment, and regrouping to derive coherent genus-level tree distributions.}
    \label{fig:framework}
\end{figure}

We approach urban tree biodiversity with an unsupervised clustering framework driven by spatial priors, using neighborhood structure and planting patterns as the primary signal. Rather than relying on datasets that label each tree by species or genus and then training a supervised model on those labels, our method operates without labels and instead groups trees into taxonomically coherent clusters using visual embeddings extracted from street-level imagery. Each cluster functions as a pseudo-taxon, allowing the computation of Shannon, Simpson, and related indices without the need for explicit species identification. The end-to-end pipeline consists of five stages: data description, spatial and visual embedding extraction, spatial–visual clustering, cluster refinement, and biodiversity metric computation. An overview of the methodology is shown in Fig. \ref{fig:framework} and detailed later in this section.

\subsection{Data Description}\label{Data}
This study relies on the Auto Arborist dataset \cite{beery_auto_2022}, a large-scale benchmark of urban street trees derived from dense sampling of Google Street View (GSV) imagery. While the dataset spans numerous cities, we selected the eight most diverse in terms of urban form, genera composition, and spatial distribution to provide a robust testbed for our clustering framework. Across these cities, the dataset contains hundreds of thousands of geotagged tree instances annotated at the genus level. The dataset is no longer publicly hosted due to maintenance constraints, but remains available from the authors upon request. Full details of the acquisition and preprocessing pipeline are provided in \cite{beery_auto_2022}. To ensure reproducibility and enable further research, we will release the spatial and visual embeddings of some cities generated in this study, providing a standardized, high-quality resource for consistent and cross-city evaluation.

\begin{table}[h]
\caption{Summary of urban forest characteristics for the eight North American cities in the Auto Arborist dataset. Urban area, total tree counts, number of genera, genera imbalance (mean $\pm$ standard deviation of the max/min ratio), and maximum observed Shannon diversity ($H'_{\text{max}}$) are reported.}
\label{tab:data_summary}
\centering
\begin{tabular}{@{}lccccc@{}}
\toprule
City & Urban Area (km$^2$) & \#Trees & \#Genera & $\mu_{\mathrm{IR}} \pm \sigma_{\mathrm{IR}}$ & $H'_{\text{max}}$ \\
\midrule
Calgary, AB        & 621.7   & 64,476   & 35  & 11.28 $\pm$ 18.02  & 2.47 \\
New York, NY       & 1,223.6 & 560,069  & 68  & 11.33 $\pm$ 14.67  & 3.12 \\
Columbus, OH       & 1,336.5 & 114,536  & 81  & 17.96 $\pm$ 24.95  & 3.18 \\
Denver, CO         & 1,760.0 & 175,438  & 97  & 18.40 $\pm$ 20.25  & 3.27 \\
Los Angeles, CA    & 5,910.0 & 391,788  & 202 & 20.98 $\pm$ 27.90  & 3.50 \\
Seattle, WA        & 2,546.0 & 150,983  & 142 & 34.06 $\pm$ 37.07  & 3.44 \\
Washington, DC     & 3,351.0 & 152,983  & 71  & 43.13 $\pm$ 43.38  & 2.89 \\
San Francisco, CA  & 1,330.0 & 154,698  & 195 & 57.96 $\pm$ 77.78  & 3.48 \\
\midrule
\textbf{Average}   & 2,384.5 & 220,746  & 111 & 26.43 $\pm$ 32.00  & 3.17 \\
\botrule
\end{tabular}
\end{table}

The dataset used encompasses eight major North American cities (Table~\ref{tab:data_summary}), capturing substantial variation in urban extent, tree abundance, taxonomic richness, and genera imbalance. Urban area spans nearly an order of magnitude, from 621.7 $km^2$ in Calgary to 5,910 $km^2$ in Los Angeles. Total tree counts range from roughly 64,000 in Calgary to over 560,000 in New York, reflecting differences in both city size and canopy coverage. Taxonomic richness, measured as the number of genera, peaks in Los Angeles (202) and San Francisco (195), indicating broad genera variety, and is lowest in Calgary (35), suggesting a more homogeneous composition.

The genera imbalance factor (IR), defined as the mean $\pm$ standard deviation of the maximum-to-minimum (MM) genus count ratio within city grids, highlights marked contrasts in dominance patterns. New York exhibits the most even composition (11.33 $\pm$ 14.67), whereas San Francisco (57.96 $\pm$ 77.78), Washington (43.13 $\pm$ 43.38), and Seattle (34.06 $\pm$ 37.07) display strong skew toward a few dominant genera. Such high imbalance factors present greater challenges for biodiversity estimation, particularly for richness metrics. Importantly, IR should be interpreted as a measure of dominance rather than a direct indicator of overall biodiversity: cities with high IR can still host many genera, but those genera are unevenly distributed, with a few disproportionately represented. In our analysis, we found that while extreme IR values often coincided with higher richness errors, they did not consistently predict lower Shannon or Simpson diversity, since these indices emphasize evenness and relative abundance rather than absolute richness.

Shannon diversity, which in this dataset can reach values of approximately 3.5, further distinguishes the cities. Larger and compositionally diverse cities such as Los Angeles, San Francisco, and Seattle attain the highest maxima, reflecting both high richness and evenness in certain localities. By contrast, cities with fewer genera or pronounced dominance, such as Calgary and Washington, tend toward lower Shannon maxima, consistent with reduced local evenness.

Collectively, these differences position the dataset used as a diverse and challenging benchmark to test the generalizability of unsupervised clustering methods in urban biodiversity assessment. A detailed pseudo-code of the pipeline is provided in Appendix~\ref{ablation} (Algorithm \ref{alg:pipeline}), and the complete implementation, including hyperparameter tuning and preprocessing scripts, is available on Github.
% \begin{figure}[h]
%     \centering
%     \includegraphics[width=\linewidth]{Figures/Spatial_Embeddings.png}
%     \caption{Generation of pseudo-clusters using spatial priors. Street-level images are sampled within local grids, visual embeddings are extracted, and spatial proximity guides grouping into initial clusters.}
    
%     \label{fig:Step1}
% \end{figure}
\subsection{Clustering Framework}
Clustering urban tree data from street-level imagery requires methods that can accommodate both fine-grained visual differences between genera and spatial patterns imposed by urban design. Our framework combines visual and spatial embeddings in a multi-stage pipeline that progressively refines initial groupings, removing visual outliers, merging highly similar clusters, and reassigning ambiguous samples. This design balances taxonomic accuracy with spatial representativeness, ensuring that diversity metrics capture both the magnitude and spatial structure of urban forest biodiversity.
\subsubsection{Pseudo Clustering}\label{sec23}
Urban trees are often arranged linearly along streets and boulevards to provide shade, enhance aesthetic appeal, and define pedestrian zones \cite{mattocks_street_1924}. However, many clustering algorithms perform poorly on linear or anisotropic spatial patterns, as they often assume compact clusters with uniform densities. Clustering methods like Ordering Point to Identify the Clustering Structure (OPTICS) \cite{ankerst_ordering_2008} or Fuzzy c-Means (FCM) \cite{bezdek_fcm_1984} rely on parameters or assumptions that are not well suited for elongated or non-convex cluster geometries.

Spatial embeddings derived from TaxaBind \cite{sastry2024taxabindunifiedembeddingspace}, a multimodal foundation model that incorporates geographic location and environmental features globally, provide a powerful solution to the limitations of traditional clustering algorithms when applied to linearly arranged urban tree data. Unlike centroid- or density-based methods, which often fail to capture elongated or irregular structures \cite{bhattacharjee2021survey}, spatial embeddings transforms raw geographic coordinates into a learned representation that encodes relative spatial context and structural relationships. This allows linearly aligned trees to be represented as coherent patterns, even in the presence of spacing variability, gaps, or missing data. By preserving and enhancing neighborhood relationships, spatial embeddings help group trees that share local spatial context, making clustering more robust to urban morphology.

For visual features, we use BioCLIP \cite{stevens_bioclip_2024} to extract fine-grained, taxonomically relevant representations from street-level imagery. These visual embeddings complement the spatial representation by capturing leaf, bark, and crown characteristics that are essential for distinguishing between visually similar genera.

As an initial step, we group TaxaBind spatial embeddings into pseudo-clusters using the HDBSCAN algorithm \cite{Malzer_2020}, leveraging the tendency of urban trees to be planted in structured, often linear arrangements. This approach typically results in taxonomically pure clusters due to the spatial regularity introduced by common urban planting strategies such as row planting along streets or uniform spacing in public parks.
\begin{figure}
\centering
\includegraphics[width=0.8\linewidth]{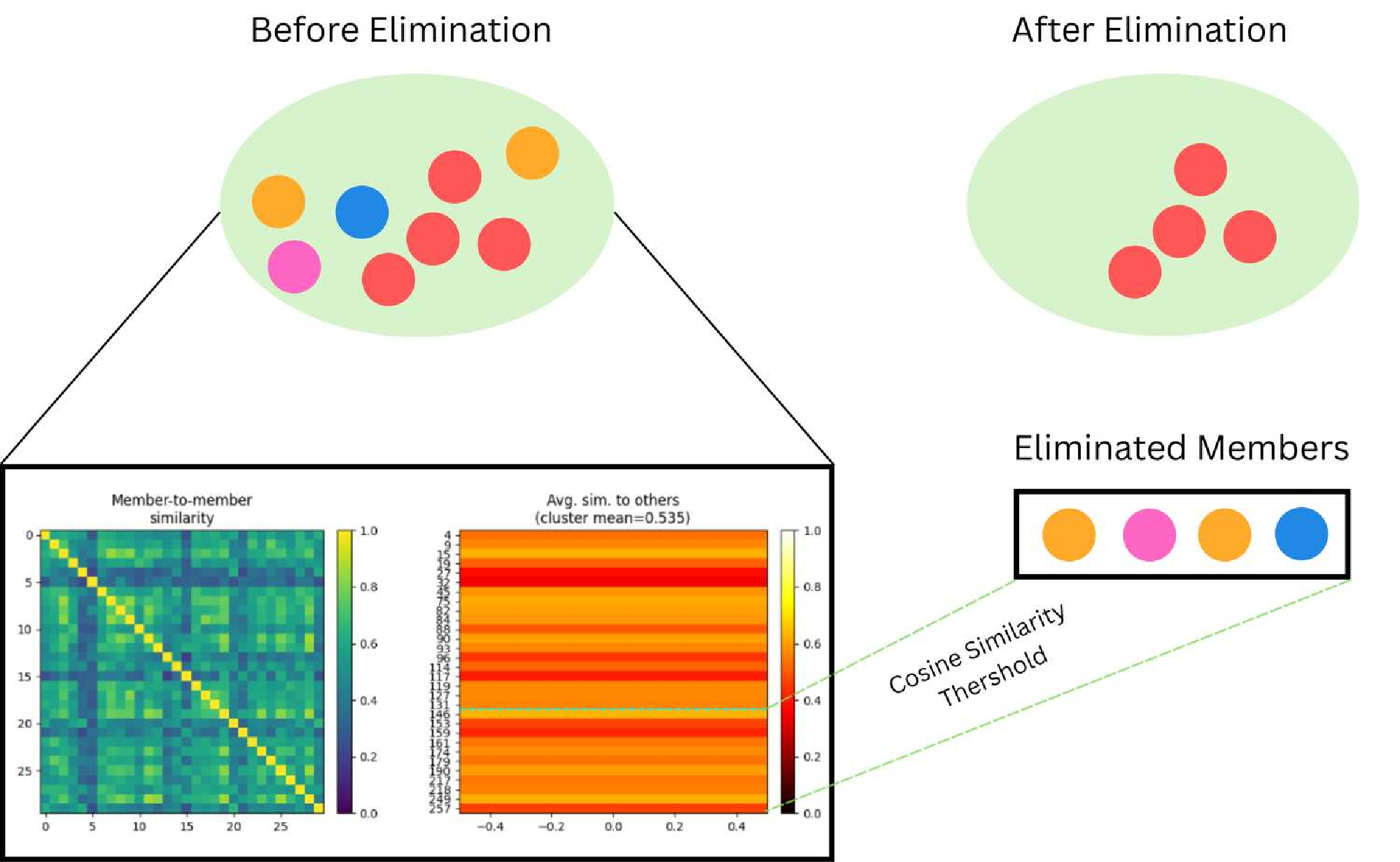}
\caption{Intra-cluster similarity filtering using a tuned threshold. Low-similarity members (highlighted) are excluded to increase cluster coherence.}
\label{fig:Step2}
\end{figure}
\subsubsection{Outlier Elimination}
Outlier elimination based on intra-cluster visual similarities aims to improve the coherence and reliability of clustering by removing samples that visually deviate from the dominant patterns within their assigned group. After forming initial clusters based on spatial proximity, we compute pairwise cosine similarities \cite{xia2015learning} between all visual embeddings within each cluster. Cosine similarity is particularly suitable here because it measures the angular closeness of vectors rather than their raw magnitudes, which is advantageous in high-dimensional embedding spaces where differences in vector length are less informative than differences in orientation. This makes it effective for capturing whether samples share a common visual representation, even when the embedding space is large and sparse. Samples with consistently low similarity to other members are considered visually inconsistent and are eliminated as outliers.

Because visual differences among tree genera can be subtle and fine-grained, especially in urban settings with morphologically similar genera, directly detecting outliers is challenging. To address this, we adopt a conservative strategy in which outliers are identified as those samples whose average similarity to other cluster members falls below a threshold determined through hyperparameter tuning (see Appendix~\ref{appendix:hyperparam}). This thresholded filtering prioritizes consistency while avoiding premature exclusion of valid but visually nuanced samples, resulting in more stable and interpretable clusters in subsequent steps. This process is further illustrated in Fig. \ref{fig:Step2}.

\subsubsection{Re-grouping and Re-assignment}
The re-grouping and re-assignment process targets samples eliminated during the outlier detection stage, which were initially clustered based primarily on spatial proximity rather than visual coherence. Since such pseudo-clusters may group together visually dissimilar trees due to their physical arrangement, the eliminated members are revisited to assess whether they form more semantically consistent groupings.

Visual embeddings of all eliminated samples are compared, and if a subset exhibits mutual similarity above the tuned cosine similarity threshold (Appendix~\ref{appendix:hyperparam}), eliminated samples are grouped into a new cluster, capturing visual structure that spatial proximity alone may have missed. Remaining samples are evaluated for reassignment to the most visually similar existing cluster, again using the tuned threshold. If no such reassignment is possible, the sample is treated as a singleton cluster.

This step enhances visual coherence in the final clustering while improving the model’s ability to capture the long-tail distribution of urban tree genera where many rare or underrepresented classes naturally appear as singletons.

% \begin{figure}
% \centering
% \includegraphics[width=0.8\linewidth]{Figures/Reassignment.png}
% \caption{Outliers are either re-grouped into new visually coherent clusters or reassigned to existing ones, using tuned similarity thresholds.}
% \label{fig:Step3}
% \end{figure}

\subsubsection{Merging Process}
\begin{figure}
    \centering
    \includegraphics[width=0.7\linewidth]{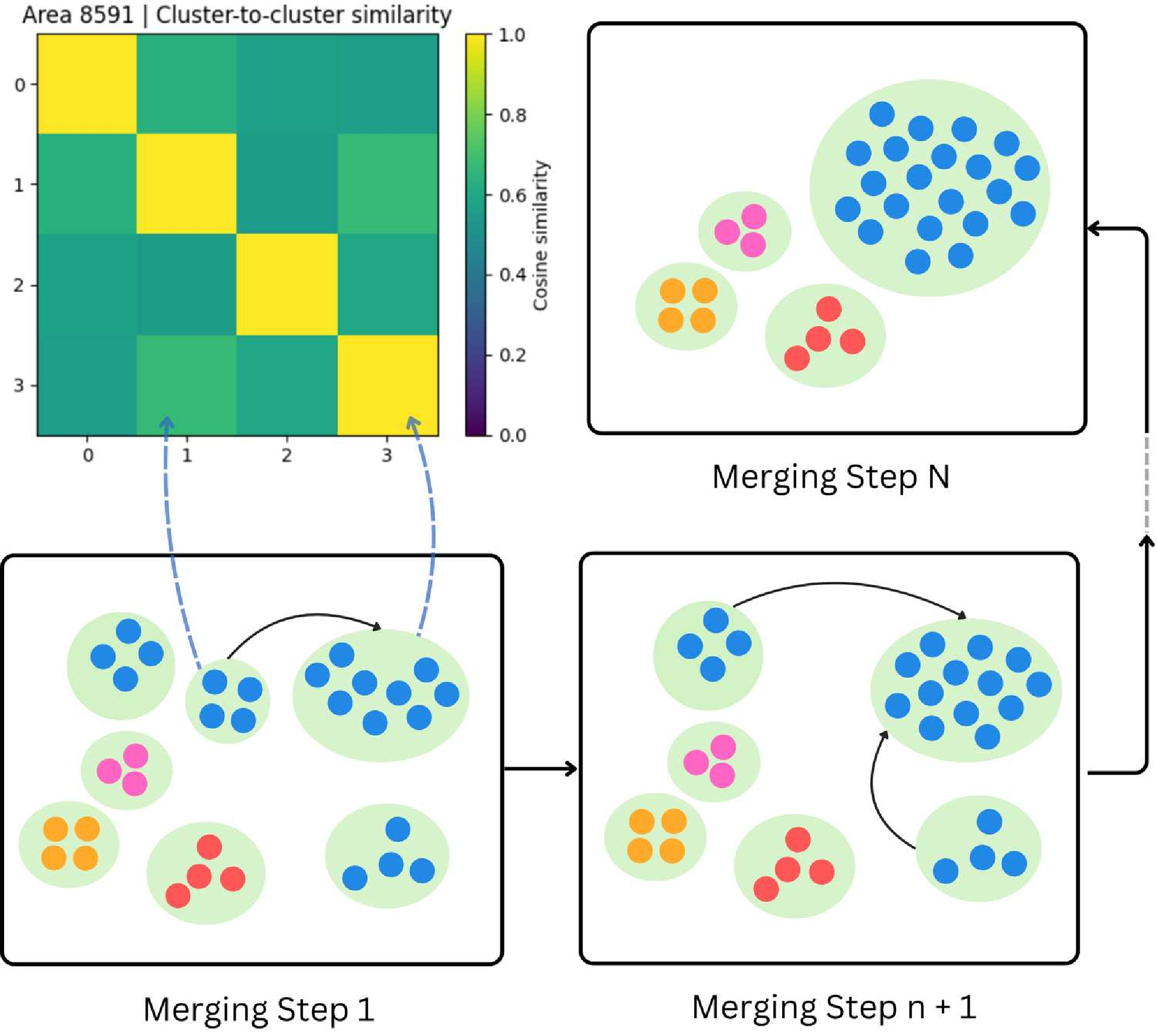}
    \caption{ Iterative merging of visually similar clusters, guided by centroid similarity, to consolidate over-fragmented groups.}
    \label{fig:Step4}
\end{figure}

Within each post-processing iteration, cluster merging serves as an adaptive consolidation step, reducing over-fragmentation as the clustering structure evolves. After outlier elimination, some clusters become more visually coherent, and new clusters may emerge from outlier grouping. These updates can reveal opportunities to combine clusters that now exhibit strong visual similarity.

As shown in Fig. \ref{fig:Step4}, we compute the average cosine similarity between the embedding centroids of all cluster pairs. If the similarity between two clusters exceeds a threshold determined through hyperparameter tuning (see Appendix~\ref{appendix:hyperparam}), they are merged into a single group. Performing this merging iteratively, rather than in a single final pass, ensures that decisions are made on the most up-to-date cluster compositions, accounting for changes introduced by outlier removal, grouping, and reassignment.

This incremental approach prevents premature large-scale merges that risk collapsing meaningful fine-grained distinctions, while still allowing the framework to consolidate redundant clusters as they appear. Over successive iterations, the result is a set of visually coherent, non-redundant clusters that better reflect the ecological structure of the dataset. By reducing spurious fragmentation and redundancy, these refinements directly improve the accuracy of abundance distributions, which in turn leads to more reliable Shannon and Simpson diversity estimates.

\bigskip
% \begin{minipage}{\hsize}%
% \lstset{frame=single,framexleftmargin=-1pt,framexrightmargin=-17pt,framesep=12pt,linewidth=0.98\textwidth,language=pascal}% Set your language (you can change the language for each code-block optionally)
% %%% Start your code-block
% \begin{lstlisting}
% for i:=maxint to 0 do
% begin
% { do nothing }
% end;
% Write('Case insensitive ');
% Write('Pascal keywords.');
% \end{lstlisting}
% \end{minipage}

\section{Conclusion}\label{sec13}

This study introduced an unsupervised clustering framework for quantifying urban forest diversity directly from street level imagery, capturing both evenness and dominance based diversity indices. Extensive hyperparameter tuning and ablation experiments revealed that combining outlier elimination with cluster merging provided the best trade off between clustering quality, richness accuracy, and preservation of diversity structure, while additional reassignment steps offered only marginal benefits. Across diverse urban contexts, the method achieved low Wasserstein distances for both Shannon and Simpson diversity, indicating strong agreement with ground truth even in cities with highly imbalanced genera distributions.

Importantly, the framework preserved the spatial organization of diversity, as reflected by minimal differences in Moran’s I values for Shannon and Simpson indices across cities. This ability to recover both the magnitude and spatial structure of key biodiversity metrics underscores the method’s potential as a scalable monitoring tool for urban tree communities. While challenges remain including uneven imagery availability, occlusion from privacy blurring, and environmental variability, these results demonstrate the promise of computer vision based approaches for large scale, fine grained biodiversity assessment and their potential integration into urban forestry policy and planning.

\backmatter

\bmhead{Acknowledgements}
The authors thank Dubai Future Foundation, UnipolTech, Consiglio per la Ricerca in Agricoltura e l’Analisi dell’Economia Agraria, Volkswagen Group America, FAE Technology, Samoo Architects \& Engineers, Shell, GoAigua, ENEL Foundation, Kyoto University, Weizmann Institute of Science, Universidad Autónoma de Occidente, Instituto Politecnico Nacional, Imperial College London, Universitá di Pisa, KTH Royal Institute of Technology, AMS Institute and all the members of the MIT Senseable City Lab Consortium for supporting this research. 

\section*{Declarations}

% Some journals require declarations to be submitted in a standardised format. Please check the Instructions for Authors of the journal to which you are submitting to see if you need to complete this section. If yes, your manuscript must contain the following sections under the heading `Declarations':

% \subsection{Funding}
% This work was supported by the  project.

\subsection{Conflict of interest}
The authors declare no competing interests.

\subsection{Ethics approval and consent to participate}
Not applicable.

\subsection{Consent for publication}
Not applicable.

\subsection*{Data availability}
The visual embeddings derived from BioCLIP for one representative city, along with all processed spatial embeddings from TaxaBind, are available from the corresponding author upon request.

\subsection{Materials availability}
Not applicable.

\subsection{Code availability}
The code used for analysis is available at \url{https://github.com/Diaa340/Releaf/}.

\subsection{Author contribution}
M.M., F.D., I.Z., M.S., and D.A.A. conceptualized the study. D.A.A. and M.S. performed the methodology development and data curation, conducted the formal analysis, and wrote the original draft. M.M., F.D., and I.Z. supervised the research, C.R. provided resources, and handled funding acquisition. All authors reviewed and edited the manuscript. 

% \noindent
% If any of the sections are not relevant to your manuscript, please include the heading and write `Not applicable' for that section. 

\bigskip

% Springer journals and proceedings: \url{https://www.springer.com/gp/editorial-policies}

% Nature Portfolio journals: \url{https://www.nature.com/nature-research/editorial-policies}

% \textit{Scientific Reports}: \url{https://www.nature.com/srep/journal-policies/editorial-policies}

% BMC journals: \url{https://www.biomedcentral.com/getpublished/editorial-policies}
% \end{flushleft}

\begin{appendices}

\appendix

\section{Hyperparameter Tuning}\label{appendix:hyperparam}

We optimized the clustering pipeline using Latin hypercube sampling (LHS) \cite{shields_150706716_2016} to efficiently explore the multidimensional parameter space (Table~\ref{tab:hyperparams}). The LHS sample size was set to $n=100$, following established guidelines of selecting $n \approx 10$--$15$ times the number of tuned parameters \cite{loeppky2009choosing}, which for our 7 parameters (outlier fraction, minimum cluster size, minimum samples, similarity thresholds for merging, outlier grouping, and reassignment, and the number of post-processing iterations) ensures systematic space-filling coverage without the redundancy of exhaustive grid search.

The optimal configuration (Table \ref{tab:hyperparams}) favored lenient clustering thresholds (\texttt{min\_cluster\_size} = 2, \texttt{min\_sample} = 3), increasing spatial coverage by allowing clusters to form from few points. A relatively high number of post-processing iterations (\texttt{N\_iter} = 5) compensated for the over-fragmentation that can occur under these settings. The merging threshold (0.68) lay toward the upper end of its range, ensuring that only visually very similar clusters were combined, while thresholds for grouping and reassigning outliers were low-to-moderate, preventing overly aggressive merges that could reduce taxonomic purity. The selected outlier fraction (0.16) balanced removal of inconsistent samples with retention of rare, ecologically important cases.

The performance of each configuration was assessed using a weighted score designed to capture clustering quality, biodiversity accuracy, and spatial applicability:

\begin{equation}
\text{Score} = 0.3V' + 0.3\text{RMSE}' + 0.3S' + 0.1C'
\end{equation}

where $V'$, $\text{RMSE}'$, $S'$, and $C'$ denote the normalized V-score, RMSE of richness, Shannon Wasserstein distance, and coverage, respectively. Equal weights were assigned to the first three terms to prevent any single metric from dominating and to ensure balanced optimisation of taxonomic alignment, richness accuracy, and diversity preservation. Shannon diversity was selected over Simpson diversity due to their high correlation naturally, which could bias results if both were included. Coverage was given a smaller weight (0.1) but retained in the score because stricter clustering parameters can markedly reduce the number of spatial units with valid biodiversity estimates; incorporating coverage penalizes configurations that, while accurate, apply to only a limited portion of the study area.

\begin{table}[h]
\caption{Hyperparameter search space and optimal configuration from LHS tuning.}
\label{tab:hyperparams}
\centering
\begin{tabular}{@{}lcc@{}}
\toprule
Hyperparameter & Range & Best value \\
\midrule
outlier\_frac         & 0.1--0.3 & 0.16 \\
min\_cluster\_size    & 2--10    & 2 \\
min\_samples          & 1--10    & 3 \\
sim\_thresh\_merge    & 0.5--0.9 & 0.68 \\
sim\_thresh\_outliers & 0.5--0.9 & 0.53 \\
sim\_thresh\_reassign & 0.5--0.9 & 0.57 \\
N\_iter               & 0--6     & 5 \\
\botrule
\end{tabular}
\end{table}

\begin{algorithm}[H]
\caption{Unsupervised Clustering Pipeline}
\label{alg:pipeline}
\begin{algorithmic}[1]
\Require Area image embeddings $\mathbf{X}_{\text{img}}$, location embeddings $\mathbf{X}_{\text{loc}}$
\Statex \textbf{Step 1: Initial clustering}
\State Compute cosine distance matrix $D$ for $\mathbf{X}_{\text{loc}}$.
\State Apply HDBSCAN to $D$ with parameters \texttt{min\_cluster\_size}, \texttt{min\_samples} to obtain initial labels $\mathbf{z}$.

\Statex \textbf{Step 2: Iterative post-processing on $\mathbf{X}_{\text{img}}$}
\For{$t \gets 1$ \textbf{to} \texttt{N\_iter}}
  \State \textbf{(a) Outlier elimination:}
  \State For each cluster, compute mean intra-cluster similarity.
  \State Mark the lowest \texttt{outlier\_frac} fraction as outliers ($z_i \gets -1$).

  \State \textbf{(b) Cluster merging:}
  \State Merge clusters whose mean embedding similarity $> \texttt{sim\_thresh\_merge}$.

  \State \textbf{(c) Outlier grouping:}
  \State Group outliers whose mutual similarity $> \texttt{sim\_thresh\_outliers}$ into new clusters.

  \State \textbf{(d) Outlier reassignment:}
  \State Reassign remaining outliers to the closest cluster if similarity $> \texttt{sim\_thresh\_reassign}$.

  \State Update labels $\mathbf{z}$.
\EndFor

\State \Return Final cluster labels $\mathbf{z}$
\end{algorithmic}
\end{algorithm}

\section{Ablation analysis}\label{ablation}

We evaluated the contribution of each post-processing stage by selectively enabling or disabling outlier elimination, merging, and reassignment (Algorithm \ref{alg:pipeline}). 

Performance was ranked using the same weighted score as in tuning, with higher weight on Shannon Wasserstein distance to emphasize diversity preservation as seen in Table \ref{tab:ablation}.

\begin{table}[h]
\caption{Ablation study results.}
\label{tab:ablation}
\centering
\begin{tabular}{@{}ccccccc@{}}
\toprule
Elim & Merge & Reassign & V-score $\uparrow$ & $RMSE_{\alpha} \downarrow$ & $W_{\text{Shannon}} \downarrow$ & Score $\uparrow$ \\
\midrule
$\checkmark$ & $\checkmark$ &              & 0.466 &  5.54 & 0.524 & 0.881 \\
             & $\checkmark$ & $\checkmark$ & 0.565 &  5.70 & 0.615 & 0.865 \\
             & $\checkmark$ &              & 0.583 &  5.38 & 0.602 & 0.857 \\
$\checkmark$ & $\checkmark$ & $\checkmark$ & 0.479 &  8.13 & 0.698 & 0.789 \\
$\checkmark$ &              &              & 0.553 & 14.41 & 0.769 & 0.681 \\
             &              &              & 0.675 & 14.31 & 0.907 & 0.660 \\
             &              & $\checkmark$ & 0.659 & 14.79 & 0.922 & 0.647 \\
$\checkmark$ &              & $\checkmark$ & 0.560 & 17.66 & 1.018 & 0.581 \\
\bottomrule
\end{tabular}
\end{table}

The combination of elimination and merging provided the best overall trade-off between taxonomic coherence, richness accuracy and spatial diversity structure, despite not producing the highest V-score. Merging alone increased V-score but reduced diversity preservation, while reassignment offered limited gains and sometimes increased richness error. Removing merging entirely caused substantial degradation, confirming its central role in counteracting over-fragmentation. Elimination without merging improved internal consistency but failed to recover broader genera  groupings. These results show that the two-stage refinement of elimination followed by targeted merging is essential for maintaining both the magnitude and spatial organization of biodiversity indices.

% \begin{figure}
%     \centering
%     \includegraphics[width=\linewidth]{Figures/GradCAM.png}
%     \caption{Input images in (a), (b), (c), and (d) along with the extracted GradCAM maps of BioCLIP Embeddings in (e), (f), (g) and (h) respectively.}
%     \label{fig:CAM}
% \end{figure}

\section{Spatial fidelity results.}\label{spatial}
Spatial pattern fidelity was assessed using Moran’s I (Table \ref{tab:morans_pred}). Across all cities and diversity metrics, predictions exhibited strong and highly significant spatial autocorrelation ($p<0.05$). Average Moran’s I values were 0.653 for richness, 0.729 for Shannon, and 0.714 for Simpson, with consistently high scores across cities. 
These results indicate that the predicted diversity maps not only preserve spatial clustering but also reproduce city-level patterns of aggregation with high fidelity. Notably, richness showed the greatest variation between cities (e.g., Denver = 0.823, Los Angeles = 0.524), whereas Shannon and Simpson displayed more uniformly high clustering across contexts. 

\begin{table}[h]
\caption{Spatial autocorrelation (Moran's I) of predictions. All values significant at $p<0.05$.}
\label{tab:morans_pred}
\centering
\begin{tabular}{lccc}
\toprule
City & Richness & Shannon & Simpson \\
\midrule
Calgary        & 0.549 & 0.617 & 0.612 \\
New York       & 0.713 & 0.766 & 0.769 \\
Columbus       & 0.577 & 0.652 & 0.633 \\
Denver         & 0.823 & 0.860 & 0.836 \\
Los Angeles    & 0.524 & 0.566 & 0.519 \\
Seattle        & 0.683 & 0.783 & 0.764 \\
Washington     & 0.726 & 0.823 & 0.816 \\
San Francisco  & 0.630 & 0.761 & 0.762 \\
\midrule
\textbf{Average} & \textbf{0.653} & \textbf{0.729} & \textbf{0.714} \\
\bottomrule
\end{tabular}
\end{table}

\section{Supervised Diagnostics}\label{diagnosis}

\begin{table}[h]
\centering
\caption{Linear probing results (supervised baseline). Reported are RMSE for species richness and Wasserstein distances ($W_1$) for Shannon and Simpson indices.}
\label{tab:linear_probing}
\begin{tabular}{lccc}
\toprule
City & $RMSE_{\alpha}$ $\uparrow$ & $W_1$--Shannon $\downarrow$ & $W_1$--Simpson $\downarrow$  \\
\midrule
Calgary         & 1.91 & 0.306 & 0.140 \\
New York        & 2.62 & 0.283 & 0.114 \\
Columbus        & 3.25 & 0.405 & 0.143 \\
Denver          & 3.40 & 0.320 & 0.093 \\
Los Angeles     & 6.75 & 0.644 & 0.157 \\
Seattle         & 2.47 & 0.413 & 0.132 \\
Washington, D.C.& 3.48 & 0.299 & 0.092 \\
San Francisco   & 6.82 & 0.361 & 0.062 \\
\bottomrule
\end{tabular}
\end{table}

\noindent
Linear probing provides a strong supervised baseline, yielding low richness RMSE across most cities. However, Wasserstein distances for Shannon and Simpson distributions remain relatively high, particularly in more diverse settings such as Los Angeles and San Francisco. This indicates that while supervised embeddings are precise for richness prediction, they do not fully capture the community structure of urban biodiversity.

\medskip

Tables~\ref{tab:inflation_diag_clustering} and \ref{tab:inflation_diag_dist} report diagnostic indicators for the unsupervised clustering pipeline. Were the reduced Wasserstein distances merely the result of entropy inflation, we would expect pronounced oversplitting, the proliferation of singleton clusters, substantial positive shifts in evenness, and weak rank--abundance correlations. Instead, across all cities the singleton ratio was zero, oversplitting factors remained close to one, shifts in evenness were small (including a slight decrease in New York), and rank--abundance correlations were consistently high ($\rho \approx 0.9$). These results indicate that the unsupervised approach preserves the ecological structure of community distributions, rather than achieving lower errors through artificial entropy inflation.

\begin{table}[h]
\centering
\caption{Clustering diagnostics for the unsupervised pipeline. Oversplitting factor close to 1, high purity, and $\Delta$Evenness near 0 indicate stability rather than entropy inflation.}
\label{tab:inflation_diag_clustering}
\begin{tabular}{lccc}
\toprule
City & Oversplit Factor $\approx 1$ & Purity (micro $\uparrow$) & $\Delta$Evenness $\approx 0$ \\
\midrule
Calgary         & 1.32 & 0.744 & 0.098  \\
New York        & 0.70 & 0.582 & $-0.027$ \\
Columbus        & 1.47 & 0.713 & 0.141  \\
Denver          & 1.14 & 0.610 & 0.036  \\
Los Angeles     & 1.03 & 0.641 & 0.076  \\
Seattle         & 1.73 & 0.661 & 0.103  \\
Washington, D.C.& 1.55 & 0.578 & 0.046  \\
San Francisco   & 1.13 & 0.555 & 0.039  \\
\bottomrule
\end{tabular}
\end{table}

\begin{table}[h]

\centering

\caption{Distributional fidelity of the unsupervised pipeline. High $\rho$ values and modest Shannon/Simpson gaps indicate preservation of rank--abundance structure rather than entropy inflation.}
\label{tab:inflation_diag_dist}
\begin{tabular}{lccc}
\toprule
City & $\rho$ (Rank Shape $\uparrow$) & Shannon Gap $\downarrow$ & Simpson Gap $\downarrow$ \\
\midrule
Calgary         & 0.754 & 0.508 & 0.133 \\
New York        & 0.928 & 0.486 & 0.153 \\
Columbus        & 0.840 & 0.667 & 0.192 \\
Denver          & 0.932 & 0.376 & 0.078 \\
Los Angeles     & 0.913 & 0.438 & 0.112 \\
Seattle         & 0.900 & 0.643 & 0.124 \\
Washington, D.C.& 0.912 & 0.567 & 0.069 \\
San Francisco   & 0.940 & 0.404 & 0.071 \\
\bottomrule
\end{tabular}
\end{table}

\section{Comparisons with Statistical Baselines}\label{RichnessSD}

\paragraph{Richness.} Table~\ref{tab:richness_sd} compares model error in richness estimation (RMSE) against the standard deviation (SD) of observed richness values across city grids. Normalizing error by SD ($nRMSE$) provides a scale-aware measure of accuracy, while a simple baseline (predicting mean richness per city) serves as a reference. Results indicate that our framework substantially improves over the baseline in cities such as Denver ($nRMSE=0.67$, +0.33 vs.\ baseline), Los Angeles ($nRMSE=0.62$, +0.38), New York ($nRMSE=0.64$, +0.36), and San Francisco ($nRMSE=0.77$, +0.23). By contrast, performance degrades relative to baseline in Calgary ($nRMSE=1.36$, –0.36), Columbus ($nRMSE=1.23$, –0.23), Seattle ($nRMSE=1.15$, –0.15), and Washington\ ($nRMSE=1.70$, –0.70). These findings highlight that model success is context-dependent, performing best where richness variability is high (e.g., Denver, LA) but struggling in cities with more homogeneous or imbalanced genus distributions (e.g., Washington, Calgary).

\begin{table}[h!]
\centering
\caption{Comparison of richness estimation error (RMSE) against the standard deviation (SD) of ground truth richness across cities. $nRMSE$ denotes the normalized error ($RMSE / SD$). The baseline corresponds to predicting the mean richness per city.}
\label{tab:richness_sd}
\begin{tabular}{lccccc}
\toprule
City & SD$_{\alpha}$ & RMSE$_{Model}$ & $nRMSE_{vs\,SD}$ & RMSE$_{Baseline}$ & $\Delta$ vs. Baseline \\
\midrule
Calgary        & 3.24  & 4.39  & 1.36  & 3.24  & -0.358 \\
New York       & 5.35  & 3.42  & 0.640 & 5.35  & +0.360 \\
Columbus       & 4.86  & 5.99  & 1.23  & 4.85  & -0.234 \\
Denver         & 8.09  & 5.39  & 0.666 & 8.09  & +0.334 \\
Los Angeles    & 6.01  & 3.74  & 0.622 & 6.01  & +0.378 \\
Seattle        & 8.63  & 9.94  & 1.15  & 8.63  & -0.153 \\
Washington& 9.15  & 15.6  & 1.70  & 9.15  & -0.700 \\
San Francisco  & 13.0  & 10.0  & 0.769 & 13.0  & +0.231 \\
\bottomrule
\end{tabular}
\end{table}

\paragraph{Shannon diversity.}  
As shown in Table~\ref{tab:shannon_results}, the framework outperforms a mean baseline in most cities, achieving improvements of 56–76\% in San Francisco, Denver, Los Angeles, and New York. In these contexts, $W_1$ is also much smaller than the observed within-city variability ($W_1/SD < 0.35$), indicating that the method captures evenness-sensitive structure with high fidelity. Performance is weaker in cities with more pronounced imbalance and domain shift, such as Calgary (–6.6\% vs. baseline), yet even here the error remains within the city’s observed Shannon variability ($W_1/SD < 0.90$). Overall, these results highlight that Shannon diversity is estimated reliably across diverse contexts, with the largest gains where spatial signal and visual cues are strongest.

\begin{table}[ht]
\centering
\caption{Shannon diversity results. $W_1$ denotes the Wasserstein distance of our method, compared against the baseline error ($Baseline\_W1$) and within-city variability ($SD$). Improvements are shown relative to the baseline and as the ratio to $SD$.}
\label{tab:shannon_results}
\begin{tabular}{lccccc}
\toprule
City & $W_1$ & $Baseline\_W1$ & $SD$ & Improvement vs. Baseline (\%) & $W_1/SD$ \\
\midrule
Calgary        & 0.215 & 0.202 & 0.240 &  -6.6 & 0.898 \\
New York       & 0.073 & 0.169 & 0.211 &  56.5 & 0.349 \\
Columbus       & 0.177 & 0.192 & 0.227 &   7.4 & 0.781 \\
Denver         & 0.069 & 0.198 & 0.239 &  65.2 & 0.288 \\
Los Angeles    & 0.070 & 0.166 & 0.206 &  57.8 & 0.340 \\
Seattle        & 0.155 & 0.178 & 0.217 &  12.6 & 0.714 \\
Washington& 0.133 & 0.166 & 0.217 &  20.1 & 0.611 \\
San Francisco  & 0.034 & 0.139 & 0.182 &  75.8 & 0.185 \\
\bottomrule
\end{tabular}
\end{table}

\paragraph{Simpson diversity.}  
Table~\ref{tab:simpson_results} shows consistently strong Simpson performance across all cities, with large improvements relative to baseline (42–78\%) and very low errors compared to spatial variability ($W_1/SD$ often below 0.30). Notably, cities with high genus dominance, such as Washington and San Francisco, still show very small $W_1$ values (0.033 and 0.028, respectively), confirming that the framework robustly recovers dominant-genus patterns that drive Simpson diversity. Taken together, the two tables demonstrate complementary strengths: Simpson consistently reflects dominance structure across all settings, while Shannon shows that the framework also preserves evenness gradients in most cities.

\begin{table}[ht]
\centering
\caption{Simpson diversity results. $W_1$ denotes the Wasserstein distance of our method, compared against the baseline error ($Baseline\_W1$) and within-city variability ($SD$). Improvements are shown relative to the baseline and as the ratio to $SD$.}
\label{tab:simpson_results}
\begin{tabular}{lccccc}
\toprule
City & $W_1$ & $Baseline\_W1$ & $SD$ & Improvement vs. Baseline (\%) & $W_1/SD$ \\
\midrule
Calgary        & 0.178 & 0.248 & 0.294 &  28.2 & 0.604 \\
New York       & 0.064 & 0.165 & 0.232 &  61.5 & 0.275 \\
Columbus       & 0.140 & 0.242 & 0.294 &  42.0 & 0.477 \\
Denver         & 0.040 & 0.182 & 0.249 &  78.1 & 0.160 \\
Los Angeles    & 0.067 & 0.188 & 0.251 &  64.2 & 0.267 \\
Seattle        & 0.095 & 0.175 & 0.239 &  46.0 & 0.396 \\
Washington& 0.033 & 0.137 & 0.214 &  75.9 & 0.154 \\
San Francisco  & 0.028 & 0.112 & 0.181 &  74.6 & 0.157 \\
\bottomrule
\end{tabular}
\end{table}

\paragraph{Between-city variability.}  
Comparing across cities, we observe that Simpson diversity errors are consistently lower and less variable than Shannon errors (Table~\ref{tab:simpson_results}), underscoring the robustness of dominance-sensitive indices to geographic and compositional differences. By contrast, Shannon errors show wider variation between cities (Table~\ref{tab:shannon_results}), with the largest deviations in highly imbalanced forests such as Washington and Seattle, and the smallest in more balanced systems such as Denver and New York. This pattern indicates that while the framework generalizes well across diverse urban contexts, evenness-sensitive metrics are more sensitive to genus dominance, leading to greater between-city variability. Nevertheless, in most cases the framework substantially outperformed baselines, suggesting reliable transferability across cities with heterogeneous forest structures.

\paragraph{Visual similarity–based elimination}

Outlier elimination is a key step in our clustering pipeline, intended to remove samples that are visually inconsistent with the rest of their cluster. This is performed using visual similarity scores derived from the image embeddings rather than arbitrary thresholds or random removal. The rationale is that in fine-grained biodiversity tasks, visually inconsistent samples are more likely to be the result of occlusion, poor lighting, or misclassification, and retaining them can distort both the cluster composition and subsequent biodiversity metrics.

To verify that our similarity-based elimination is indeed beneficial, and not equivalent to simply discarding samples at random, we compare it against a random elimination strategy with the same removal fraction. The comparison uses the V-score as a measure of cluster–ground truth agreement, computed after elimination. As shown in Table \ref{tab:elim_comparison}, similarity-based elimination consistently outperforms random elimination across all cities, with an average improvement of approximately 0.10 in V-score and a standard deviation of $\pm$0.02. This demonstrates that the method selectively removes genuinely inconsistent samples rather than degrading cluster quality through indiscriminate pruning.

\begin{table}[h]
\caption{Visual similarity–based elimination vs random elimination (V-scores).}
\label{tab:elim_comparison}
\centering
\begin{tabular}{@{}lcc@{}}
\toprule
City & Visual similarity–based V-score & Random elimination V-score \\
\midrule
Calgary        & 0.20 & 0.12 \\
New York       & 0.38 & 0.26 \\
Columbus       & 0.22 & 0.14 \\
Denver         & 0.25 & 0.19 \\
Los Angeles    & 0.31 & 0.21 \\
Seattle        & 0.29 & 0.20 \\
Washington& 0.26 & 0.22 \\
San Francisco  & 0.38 & 0.29 \\
\midrule
\textbf{Average} & \textbf{0.31} & \textbf{0.21} $\pm$ \textbf{0.02} \\
\bottomrule
\end{tabular}
\end{table}

\section{Validation of extracted visual features}
\begin{figure}[h]
    \centering
    \includegraphics[width=\linewidth]{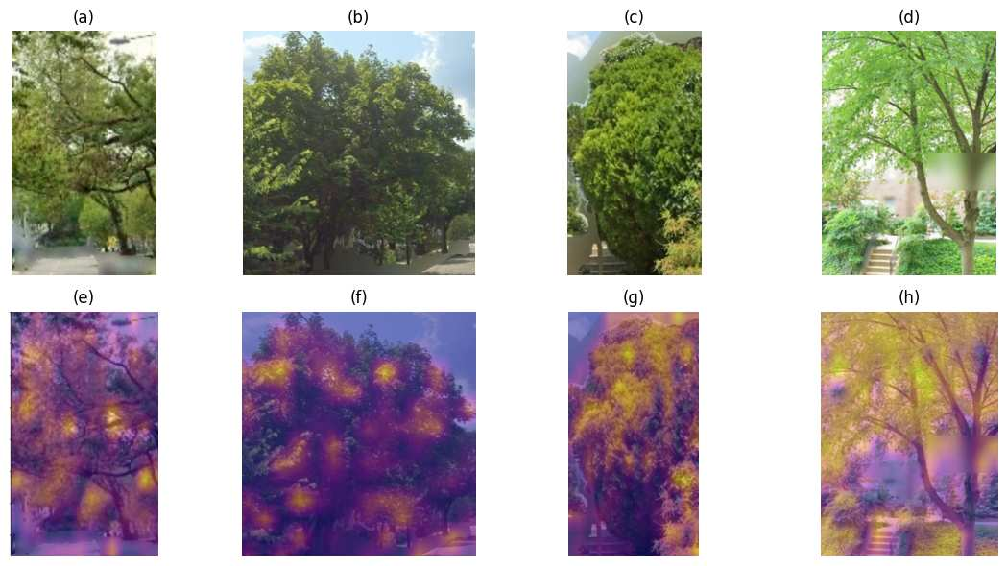}
    \caption{Input images in (a), (b), (c), and (d) along with the extracted GradCAM maps of BioCLIP Embeddings in (e), (f), (g) and (h) respectively.}
    \label{fig:CAM}
\end{figure} 
A potential concern in image-based biodiversity assessment is that the model might rely on irrelevant background elements (e.g., buildings, road markings) rather than the target object to discriminate between tree genera. Such behavior could make the visual embeddings based merging and elimination steps ineffective.

To address this, we validate the relevance of the learned visual features using Gradient-weighted Class Activation Mapping (Grad-CAM) \cite{Gradcam}. This method produces heatmaps highlighting the image regions that contribute most strongly to the network’s predictions. Visual inspection of Grad-CAM outputs confirms that the activations are concentrated on tree structures, such as leaf arrangement and branching patterns, rather than on background features (See Fig. \ref{fig:CAM}). This provides qualitative evidence that the embedding network captures biologically meaningful attributes, ensuring that downstream clustering decisions are based on ecologically relevant cues rather than scene context.




\end{appendices}


\bibliography{sn-bibliography}% common bib file

\end{document}